\documentclass[conference]{IEEEtran}
\IEEEoverridecommandlockouts

\usepackage{cite}
\usepackage{amsmath,amssymb,amsfonts}
\usepackage{graphicx}
\usepackage{textcomp}
\usepackage{xcolor}
\def\BibTeX{{\rm B\kern-.05em{\sc i\kern-.025em b}\kern-.08em
    T\kern-.1667em\lower.7ex\hbox{E}\kern-.125emX}}

\usepackage{graphicx}
\usepackage{amsmath, amsfonts, amssymb}
\usepackage{dsfont} 
\usepackage{algorithm, algpseudocode} 
\usepackage{subcaption} 
\usepackage{enumitem}
\usepackage{wrapfig}
\usepackage{pifont} 
\usepackage{stmaryrd} 
\usepackage{booktabs, multirow} 
\usepackage{tablefootnote}
\usepackage{comment}
\usepackage{colortbl} 
\usepackage[font=small]{caption}

\usepackage{xcolor}
\usepackage[colorlinks=true,linkcolor=red,citecolor=blue]{hyperref}

\newcommand{\R}{\mathds{R}}

\renewcommand{\vec}[1]{\mathbf{#1}}

\def\1{\mathbf{1}}

\def\V{\mathcal{V}}

\def\G{\mathcal{G}}
\def\H{\mathcal{H}}
\def\Prob{\mathds{P}}
\def\E{\mathds{E}}

\def\temp{\theta}

\newcommand{\etal}{\textit{et al}.\@ }
\newcommand{\eg}{\textit{e.g}.\@ }
\newcommand{\ie}{\textit{i.e}.\@ }
\newcommand{\aka}{{a.k.a}.\@ }

\definecolor{apricot}{rgb}{0.98, 0.81, 0.69}

\title{Three Bricks to Consolidate \\ Watermarks for Large Language Models
\thanks{Mail: \href{mailto:pierre.fernandez@inria.fr}{pierre.fernandez@inria.fr} - 
Code: \href{https://github.com/facebookresearch/three_bricks/}{facebookresearch/three\_bricks/}}
}

\author{
    Pierre Fernandez$^{1,3}$,
    Antoine Chaffin$^{1,2}$,
    Karim Tit$^{1}$,
    Vivien Chappelier$^{2}$,
    Teddy Furon$^{1}$
    \\  \\
    $^1$\textit{Centre Inria de l'Université de Rennes} \qquad 
    $^2$\textit{Imatag} \qquad 
    $^3$\textit{FAIR, Meta}
}
\date{}

\begin{document}

\maketitle

\begin{abstract}
Discerning between generated and natural texts is increasingly challenging. 
In this context, watermarking emerges as a promising technique for ascribing text to a specific generative model. 
It alters the sampling generation process to leave an invisible trace in the output, facilitating later detection.
This research consolidates watermarks for large language models based on three theoretical and empirical considerations. 
First, we introduce new statistical tests that offer robust theoretical guarantees which remain valid even at low false-positive rates (less than 10$^{\text{-6}}$). 
Second, we compare the effectiveness of watermarks using classical benchmarks in the field of natural language processing, gaining insights into their real-world applicability.
Third, we develop advanced detection schemes for scenarios where access to the LLM is available, as well as multi-bit watermarking. 
\end{abstract}
\vspace{0.1cm}

\vspace{0.2cm}

\section{Introduction}

The misuse of Large Language Models (LLMs) like ChatGPT~\cite{chatgpt2022}, Claude~\cite{claude}, or the open-sourced LLaMA~\cite{touvron2023llama} may become a threat as their availability and capabilities expand~\cite{weidinger2022taxonomy, crothers2022machine, cardenuto2023age}.
LLMs might help generate fake news by reducing costs to spread disinformation at scale~\cite{kertysova2018artificial, kreps2022all}, with a potential impact on public opinion and democratic outcomes~\cite{kuvsen2018politics}.
They could help impersonate people, facilitate scams~\cite{mit2023}, or make student assessments impossible.
Enforcing fair and responsible usage through regulations and technical means would be useful.

Monitoring the usage of LLMs with passive forensics is difficult because generated texts are hardly distinguishable from real ones, be it for humans or algorithms~\cite{ippolito2019automatic, mitchell2023detectgpt}.
Watermarking is a promising technique explored for generative image models~\cite{yu2022responsible, fernandez2023stable, wen2023tree} and generative text LLMs~\cite{aaronson2023watermarking,kirchenbauer2023watermark,kirchenbauer2023reliability,christ2023undetectable}.
In this case, watermarking either alters the sample generation process~\cite{aaronson2023watermarking,christ2023undetectable} or changes the probability distribution of the generated tokens~\cite{kirchenbauer2023watermark, zhao2023provable}, to leave an imperceptible trace in the generated text.
This literature then describes a detection mechanism analyzing the generated tokens to see if their distribution follows the one induced by the watermark. 

We introduce three contributions to consolidate the current literature, one for each of the following paragraphs and sections.
Each part can be read independently.

First, false positives can have serious consequences in contexts where the integrity and accuracy of results are essential, such as falsely accusing a user of producing fake news or a student of cheating in an exam.
However, current approaches~\cite{kirchenbauer2023watermark, kirchenbauer2023reliability} focus their study on sensitivity (True Positive Rate: TPR) rather than on specificity (linked to False Positive Rate: FPR).
The FPR has never been empirically checked at interesting scales (with more than 1k negative examples).
Our large-scale experiments reveal that hypotheses of previous works do not hold and that their detection thresholds largely underestimate the false positives at low FPR.
This work provides grounded statistical tests that theoretically guarantee false positive-rates and accurate p-values in real-world regimes.
We validate them empirically and show that they provide a close-to-perfect control of the FPR, even at low values ($<10^{-6}$).

Second, we compare the watermarking methods, analyzing practical implications of watermarks on traditional Natural Language Processing (NLP) benchmarks. 
Indeed, current watermark evaluation mainly considers the deviation from the original LLM distribution, for example using perplexity.
This is in contrast with the LLM litterature, where models are rather evaluated on their effective usefulness, \eg free-form completion tasks such as question answering.
Such evaluations are much more informative on the actual abilities of the model when used on downstream tasks.

Third, we expand these algorithms to advanced detection schemes.
When access to the LLM is possible at detection time, we provide optimal statistical tests.
We also investigate multi-bit watermarking (hiding binary messages as watermarks) when current approaches only tackle zero-bit watermarking.
This allows not only to determine whether the text was generated by the watermarked LLM, but also to identify which version of the model generated it.

\begin{figure}[t]
    \vspace{-0.4cm}
    \centering
    \includegraphics[width=0.95\linewidth]{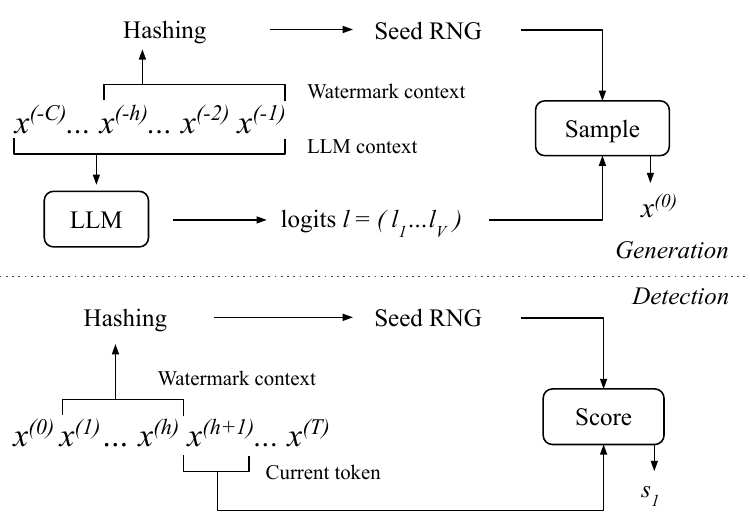}
    \caption{
    General illustration of watermarking for LLM (top: generation, bottom: detection). Details and notations in Sect.~\ref{sec:llm_wm}.}
    \label{fig:sampling}
    \vspace{-0.4cm}
\end{figure}

\section{Technical Background}

\subsection{Large Language Models (LLMs)} 

LLMs are neural networks that generate text by computing the likelihood of generating a sequence of tokens given a context~\cite{bengio2000neural}.
This paper focuses on decoder-only models, \aka auto-regressive LLMs.
The tokens are pieces of words from a vocabulary $\V$.
From a context $x^{(-C)}, ..., x^{(-1)}$, the model estimates the probability of each token of $\V$ being the next.
It computes a vector $\vec{\boldsymbol\ell}\in \R^{|\V|}$ of logits, transformed into
\begin{equation}
    \left(\Prob \left( X^{(0)}=x \big| x^{(-C)},\dots, x^{(-1)} \right)\right)_{x\in\V}=\text{softmax}(\vec{\boldsymbol\ell};\temp) 
    \label{eq:Distrib}
\end{equation}
where $\temp$ is a temperature.
The generation of a sequence from the context samples a token from this distribution, then appends it to the context and iterates the process. 
Various sampling schemes exist: greedy search, beam search, top-k sampling~\cite{fan2018hierarchical, radford2019language}, nucleus-sampling (top-p)~\cite{holtzman2019curious}, etc.

\subsection{Watermarking Text Generation}\label{sec:llm_wm}

\subsubsection{Modification of the Distribution~\cite{kirchenbauer2023reliability,kirchenbauer2023watermark, zhao2023provable}}
\label{sec:ModDistri}
The original distribution~\eqref{eq:Distrib}, denoted $\vec{p}$ for short, is replaced by a similar distribution $\vec{q} = F(\vec{p}, k)$ where $k$ is a secret key and $F$ an arbitrary function.
In the work of Kirchenbauer~\etal~\cite{kirchenbauer2023watermark}, the secret key determines a partitioning of $\V=\G_{k} \cup \bar{\G_{k}}$.
The greenlist $\G_k$ contains $\gamma |\V|$ tokens, where $\gamma \in [0,1]$.  
The logit of every token in the greenlist is incremented by $\delta>0$, and the softmax operator outputs $\vec{q}$. 
The sampling then proceeds as usual.
Intuitively, this increases the probabilty of generating greenlist tokens. 
On the other hand, $\E [F(\vec{p},K)] = \vec{p}$ so on expectation over the set of cryptographic keys, watermarking does not bias the global distribution of words ($K$ being the random variable representing the key).

The detection tokenizes the text and counts how many tokens are in their greenlist.
More formally, for a text of $T$ tokens, the score $S_T$ is the number of greenlist tokens ($x^{(t)}$ and $k^{(t)}$ respectively indicate the $t^{\textrm{th}}$ token and key):
\begin{equation}
    S_T = \sum_{t=1}^T \mathds{1} \left(x^{(t)}\in\G_{k^{(t)}}
    \right).
    \label{eq:ScoreKirchenbauer}
\end{equation}

\subsubsection{Modification of the Sampling~\cite{aaronson2023watermarking,christ2023undetectable}}
\label{sec:ModSamp}
The watermark embedding replaces the traditional sampling schemes by a deterministic process. 
For instance, Aaronson~\etal~\cite{aaronson2023watermarking} choose the next token by computing $x^{(0)} = \arg \max_{v \in \V } \vec{r}_v^{1/\vec{p}_v}$, where $\vec{p}$\footnote{(Nucleus sampling can be applied before generating $\vec{p}$)} is the distribution~\eqref{eq:Distrib} and $\vec{r}\in[0,1]^{|\V|}$ a secret vector generated from the secret key $k$. 
Intuitively, this encourages the generation of tokens that have both high $\vec{r}_v$ and $\vec{p}_v$ values.
It also presents the interesting property that $\forall v\in \V$, $\Prob (X^{(0)}=v) = \vec{p}_v$ over the randomness of the secret vector, when distributed uniformly over $[0,1]^{|\V|}$ (demonstration in App.~\ref{app:aaronson_prob}).
In other words, this watermarking does not bias the distribution on expectation over the secret vector.
 
The detection computes the following score for $T$ tokens:
\begin{equation}
S_T=-\sum_{t=1}^T \ln \left(1-\vec{r}^{(t)}_{x^{(t)}}\right).
\label{eq:ScoreAaronson}
\end{equation}

\subsection{Quality-Robustness Trade-off}\label{sec:quality-sensitivity}
For both methods we can trade off generation quality against robustness by varying the watermarking strength.
In~\cite{kirchenbauer2023watermark}, increasing the $\delta$ parameter increases the generation of green tokens at the risk of including unlikely tokens.
In~\cite{aaronson2023watermarking}, increasing the temperature $\temp$ has the same effect, since it flattens the probability vector~\eqref{eq:Distrib}, thus diminishing the relative importance of $\vec{p}_v$ over $\vec{r}_v$.

\subsection{Key Management}
\label{sec:KeyManagement}
The secret key $k$ giving birth to the greenlist $\G_k$ in~\cite{{kirchenbauer2023watermark}} or to the sampling of $\vec{r}$ in~\cite{aaronson2023watermarking} must have a wide diversity.
A fixed key causes security issues and biases the text generation.
One possibility is to make it dependent of the time $t$ as proposed in~\cite{christ2023undetectable}. 
The secret key is then different from one token to another. Yet, this brings synchronization issue at the detection stage (\eg when a sentence is deleted). 
A common practice ensuring self-synchronization - illustrated in Fig.~\ref{fig:sampling} - makes the key dependent of the window of $h$ previous tokens: $k^{(t)} = H(x^{(t-1)},\ldots,x^{(t-h)},k)$, where $H$ is a cryptographic hash function and $k$ the master key.
This secret is the seed that initializes a random number generator (RNG) at time $t$.
In turn, the RNG is used to generate the greenlist $\G_{k^{(t)}}$ or to sample $\vec{r}^{(t)}$.
The width of this window defines a trade-off between diversity of the key and robustness of the watermarking.
In the specific case where $h=0$, the key is the same for all tokens ($k^{(t)} = k$), which makes the watermarking particularly robust to text editing~\cite{zhao2023protecting}.

\subsection{$Z$-Test}
The detection tests the hypothesis $\H_0$: ``the text is natural'' (human written or written without watermark), against $\H_1$: ``the text has been generated with watermark''.

Current approaches~\cite{aaronson2023watermarking,kirchenbauer2023watermark} approximate the underlying distribution of the score $S_T$ by using a $Z$-test.
This statistical hypothesis test determines whether a sample mean differs significantly from its expectation when the standard deviation is known.
It computes the so-called $Z$ statistics:
\begin{equation}
    Z = \frac{{S_T/T - \mu_0}}{{\sigma_0 / \sqrt{T}}},
    \label{eq:Z}
\end{equation}
where \(\mu_0\) and \(\sigma_0\) are the expectation and standard deviation per token under the null hypothesis $\H_0$, \ie when the analyzed text is not watermarked.
The $Z$-test is typically used for large sample sizes assuming a normal distribution under the null hypothesis thanks to the central limit theorem. This assumption is key for computing the p-value, \ie the probability of observing a value of $Z$ at least as extreme as the one observed $z$, under the null hypothesis:
\begin{equation}
    \text{p-value}(z) = \Prob(Z>z | \H_0) = 1 - \Phi(z),
    \label{eq:pvalue}
\end{equation}
where $\Phi$ is the cumulative distribution function of the normal distribution.
At detection time, we fix a false positive rate (FPR) and flag the text as watermarked if p-value($z$) $<$ FPR.

\begin{figure*}[ht]
    \centering
    \hfill
    \begin{subfigure}{0.27\textwidth}
        \includegraphics[width=\textwidth, trim=0.2cm 0 0.3cm 0, clip]{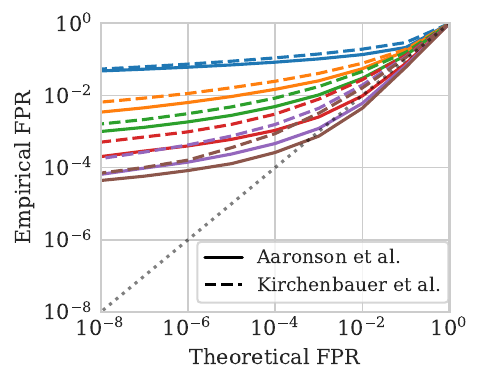}
        \caption{$Z$-tests}\label{fig:fpr-zscore}
    \end{subfigure}
    \hfill
    \begin{subfigure}{0.27\textwidth}
        \includegraphics[width=\textwidth, trim=0.2cm 0 0.3cm 0, clip]{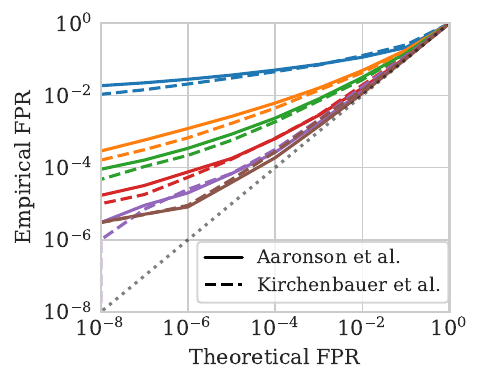}
        \caption{Tests of~\ref{sec:new-stats}}
    \end{subfigure}
    \hfill
    \begin{subfigure}{0.27\textwidth}
        \includegraphics[width=\textwidth, trim=0.2cm 0 0.3cm 0, clip]{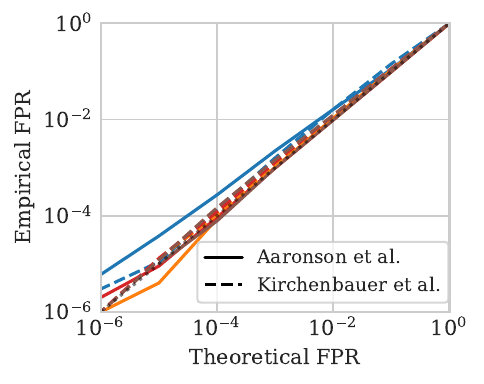}
        \caption{Tests of~\ref{sec:new-stats}, rectified with~\ref{sec:rect}}\label{fig:fpr-rect}
    \end{subfigure}
    \hfill
    \begin{subfigure}{0.125\textwidth}
        \includegraphics[width=\textwidth, trim=4.8in 0 0 0, clip]{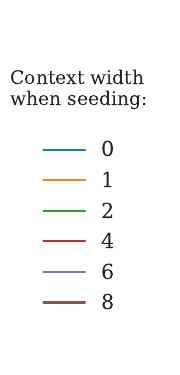}
    \end{subfigure}
    \caption{
        Empirical checks of false positive rates for different watermarks and values of the context width $h$.
        Results are computed over $10$ master keys $\times$ 100k sequences of $256$ tokens sampled from Wikipedia.
        We compare three detection tests:
        (\emph{Left}) using $Z$-tests;
        (\emph{Middle}) using new statistical tests presented in~\ref{sec:new-stats};
        (\emph{Right}) using the new statistical tests with the rectified scoring strategy of~\ref{sec:rect}.
        Theoretical values do not hold in practice for $Z$-tests, even for high values of $h$, and empirical FPRs do not match theoretical ones.
        This is solved by basing detection on grounded statistical tests and analytic p-values, as well as by revising the scoring strategy.
    }
    \label{fig:fpr-emp}\vspace{-0.4cm}
\end{figure*}

\section{Reliability of the Detection}\label{sec:stats}

In this section, large-scale evaluations of the FPR show a gap between theory and practice. 
It is closed with new statistical tests and by rectifying the scoring method.

\vspace{-0.1cm}
\subsection{Empirical Validation of FPR with Z-Scores}\label{sec:zscore}

So far, the FPR has been checked on only around $500$ negative samples~\cite{kirchenbauer2023watermark,kirchenbauer2023reliability, zhao2023provable}.
We scale this further and select $100$k texts from multilingual Wikipedia to cover the distribution of natural text.
We tokenize with LLaMA's tokenizer, and take $T=256$ tokens/text.
We run detection tests with varying window length $h$ when seeding the RNG. 
We repeat this with $10$ different master keys, which makes $1$M detection results under $\H_0$ for each method and $h$ value.
For the detection of the greenlist watermark, we use $\gamma=0.25$.

Fig.~\ref{fig:fpr-zscore} compares empirical and theoretical FPRs.
Theoretical guarantees do not hold in practice: 
the empirical FPRs are much higher than the theoretical ones.
We also observed that distributions of p-values were not uniform (which should be the case under $\H_0$).
Besides, the larger the watermarking context window $h$, the closer we get to theoretical guarantees. 
In pratice, one would need $h>>8$ to get reliable p-values, but this makes the watermarking method less robust to attacks on generated text because it hurts synchronization.

\vspace{-0.1cm}
\subsection{New Non-Asymptotical Statistical Tests}\label{sec:new-stats}
The Gaussian assumption of $Z$-tests breaks down for short or repetitive texts.
Here are non-asymptotical tests for both methods that reduce the gap between empirical and theoretical FPR, especially at low FPR values as shown in Fig.~\ref{fig:fpr-emp}.

\subsubsection{Kirchenbauer et al. \texorpdfstring{\cite{kirchenbauer2023watermark}}{}} 
Under $\H_0$, we assume that the event $x^{(t)}\in\G_{k^{(t)}}$ occurs with probability $\gamma$, and that these events are i.i.d.
Therefore, $S_T$~\eqref{eq:ScoreKirchenbauer} is distributed as a binomial of parameters $T$ and $\gamma$. Consider a text under scrutiny whose score equals $s$.
The p-value is defined as the probability of obtaining a score higher than $s$ under $\H_0$: 
\begin{equation}
    \text{p-value}(s) = \Prob(S_T>s | \H_0) = I_{\gamma}(s,T-s+1),
\end{equation}
because $S\sim\mathcal{B}(T,\gamma)$ whose c.d.f. is expressed by $I_x(a,b)$ the regularized incomplete Beta function.

\subsubsection{Aaronson et al. \texorpdfstring{\cite{aaronson2023watermarking}}{}}
Under $\H_0$, we assume that the text under scrutiny and the secret vector are independent, so that $\vec{r}_{x^{(t)}} \overset{i.i.d.}{\sim} \mathcal{U}(0,1)$. 
Therefore, $S_T$~\eqref{eq:ScoreAaronson} follows a $\Gamma(T,1)$ distribution.
The p-value associated to a score $s$ reads:
\begin{equation}
    \text{p-value}(s) = \Prob(S_T>s | \H_0) = \frac{\Gamma(T,s)}{\Gamma(T)},
\end{equation}
where $\Gamma$ is the upper incomplete gamma function.
Under $\H_1$, the score is expected to be higher as proven in App.~\ref{app:aaronson_pvalue}, so the p-value is likely to be small.

\definecolor{shadecolor}{rgb}{0.98,0.98,0.99}
\begin{figure}[b]
    \vspace{-0.5cm}
    \fcolorbox{white}{shadecolor}{
        \includegraphics[width=0.85\linewidth, trim=0 110 0 18, clip]{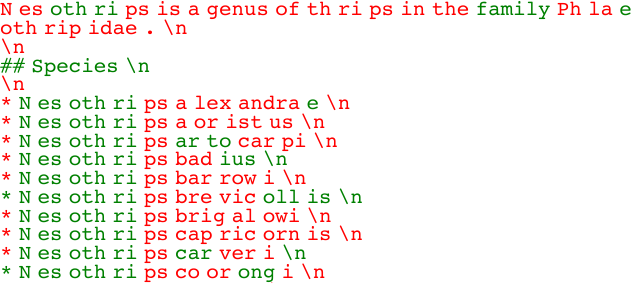}
    }
    \centering
    \caption{
    Typical example of a vanilla text with low p-value because of repeated tokens.
    It is $10^{-21}$, using the greenlist watermark with $\gamma=1/4$ and $h=2$ on $256$ tokens (we only show half of the text).
    }
    \label{fig:low-pval-text}
\end{figure}

\begin{table*}[th]
\centering
\caption{
Performances on classical free-form generation benchmarks when completion is done with watermarking.
$h$ is the watermark context width. 
We report results for methods: Aaronson~\etal\cite{aaronson2023watermarking} / Kirchenbauer~\etal\cite{kirchenbauer2023watermark}.
``-'' means no watermarking. 
}
\label{tab:bench}
\begin{tabular}{llrrrrrrr}
\toprule
 &   & GSM8K & Human Eval & MathQA & MBPP & NQ & TQA & Average \\
Model & h &  &  &  &  &  &  &  \\
\midrule
\multirow[t]{13}{*}{7B} & - & 10.3 & 12.8 & 3.0 & 18.0 & 21.7 & 56.9 & 20.5 \\
 & 1 & 10.3 / 11.1 & 12.8 / 9.8 & 2.9 / 2.8  & 18.2 / 16.0 & 21.8 / 19.5 & 56.9 / 55.3  & 20.5 / 19.1  \\
 & 4 & 10.4 / 10.8 & 12.8 / 9.2 & 3.0 / 2.8  & 17.8 / 16.4 & 21.8 / 20.2 & 56.9 / 55.1  & 20.4 / 19.1 \\
\multirow[t]{12}{*}{13B}  & - & 17.2 & 15.2 & 4.3 & 23.0 & 28.2 & 63.6 & 25.3 \\
 & 1 & 17.2 / 17.3 & 15.2 / 14.6 & 4.3 / 3.6  & 22.8 / 21.2 & 28.2 / 25.1 & 63.6 / 62.2  & 25.2 / 24.0  \\
 & 4 & 17.2 / 16.8 & 15.2 / 15.9 & 4.2 / 4.1  & 22.6 / 21.2 & 28.2 / 24.5 & 63.6 / 62.2  & 25.2 / 24.1 \\
\multirow[t]{12}{*}{30B}  & - & 35.1 & 20.1 & 6.8  & 29.8 & 33.5 & 70.0  & 32.6  \\
 & 1 & 35.3 / 35.6 & 20.7 / 20.7 & 6.9 / 7.5  & 29.6 / 28.8 & 33.5 / 31.6 & 70.0 / 69.0  & 32.7 / 32.2  \\
 & 4 & 35.1 / 34.1 & 20.1 / 22.6 & 6.9 / 7.0  & 29.8 / 28.8 & 33.5 / 31.6 & 70.0 / 68.7  & 32.6 / 32.1 \\
\bottomrule
\end{tabular}\vspace*{-0.3cm}
\end{table*}

\subsection{Rectifying the Detection Scores}\label{sec:rect}
Even with grounded statistical tests, empirical FPRs are still higher than theoretical ones.
In fact, Kirchenbauer et al.~\cite{kirchenbauer2023watermark} mention that random variables are only pseudo-random since repeated windows generate the same secret. 
This can happen even in a short text and especially in formatted data.
For instance in a bullet list, the sequence of tokens \texttt{$\backslash$n$\backslash$n*\_} repeats a lot as shown in Fig.~\ref{fig:low-pval-text}.
Repetition pulls down the assumption of independence necessary for computing the p-values.

We experimented with two simple heuristics mitigating this issue at the detection stage.
The first one takes into account a token only if the watermark context window has not already been seen during the detection.
The second scores the tokens for which the $h+1$-tuple formed by \{watermark context + current token\} has not already been seen.
Note, that the latter is present in~\cite{kirchenbauer2023watermark}, although without ablation and without being used in further experiments.
Of the two, the second one is better since it counts more ngrams, and thus has better TPR. 
It can also deal with the specific case of $h=0$.

Figure~\ref{fig:fpr-rect} reports empirical and theoretical FPRs when choosing not to score already seen $h+1$-tuples.
They now match perfectly, except for $h=0$ where the FPR is still slightly underestimated.
\emph{In short, we guarantee FPR thanks to new statistical tests and by scoring only tokens for which \{watermark context + current token\} has not been scored.}


\section{Watermark Evaluation}

This section introduces evaluation with the revised statistical tests, and investigate the impact of LLM watermarking on classical NLP benchmarks.

\subsection{Robustness Analysis}

We now compare watermarking methods by analyzing the TPR when detecting watermarked texts.
For detection, we employ the previous statistical tests and scoring strategy.
We flag a text as watermarked if its p-value is lower than $10^{-5}$ ensuring an FPR=$10^{-5}$.
For these experiments, we stay close to a chatbot scenario.
We prompt Guanaco-7b~\cite{dettmers2023qlora}, an instruction fine-tuned version of LLaMA, with the first $1$k prompts from the Alpaca dataset~\cite{alpaca}.
For generation, we use top-$p$ sampling with $p=0.95$, and in the case of \cite{kirchenbauer2023watermark} a temperature $\theta =0.8$ and $\gamma=1/4$.
We simulate synonym attacks by randomly replacing tokens with probability $0.3$ (other attacks are studied in related work~\cite{kirchenbauer2023reliability}).

Tab.~\ref{tab:robustness} reports the TPR for different strength of the watermark (see Sect.~\ref{sec:quality-sensitivity}), and the S-BERT~\cite{reimers2019sentence} similarity score between the generated texts with and without watermarking to measure the semantic distortion induced by the watermark. 
Results in Tab.~\ref{tab:robustness} reveals different behaviors.
For instance, \cite{kirchenbauer2023watermark} has a finer control over the trade-off between watermark strength and quality.
Its TPR values ranges from 0.0 to 0.9, while \cite{aaronson2023watermarking} is more consistent but fails to achieve TPR higher than 0.8 even when the S-BERT score is degraded a lot.

The watermark context width also has a big influence. 
When $h$ is low, we observed that repetitions happen more often because the generation is easily biased towards certain repetitions of tokens.
It leads to average S-BERT scores below 0.5 and unusable completions.
On the other hand, low $h$ also makes the watermark more robust, especially for \cite{kirchenbauer2023watermark}.
It is also important to note that $h$ has an influence on the number of analyzed tokens since we only score tokens for which the $h+1$-tuple has not been seen before (see Sect.~\ref{sec:rect}).
If $h$ is high, almost all these tuples are new, while if $h$ is low, the chance of repeated tuples increases.
For instance in our case, the average number of scored tokens is around 100 for $h=0$, and 150 for $h=1$ and $h=4$.

\begin{table}[t]
    \centering
    \caption{
    Robustness analysis of the watermarks, with rectified statistical tests.
    We report the TPR@FPR=$10^{-5}$ and the S-BERT scores over $10\times 1$k completions, for different hyperparameters controling the strength of the watermark 
    ($\delta$ in \cite{kirchenbauer2023watermark} and $\temp$ in \cite{aaronson2023watermarking} - see Sect.~\ref{sec:quality-sensitivity}).
    The `TPR aug.' is the TPR when texts are attacked before detection by randomly replacing tokens with probability 0.3.
    }
    \label{tab:robustness}
    \resizebox{\linewidth}{!}{
        \begin{tabular}{rl @{\hspace{0.3cm}} rrrr @{\hspace{0.5cm}} rrrr}
        \toprule
        & & \multicolumn{4}{c}{Aaronson~\etal\cite{aaronson2023watermarking}} &  \multicolumn{4}{c}{Kirchenbauer~\etal\cite{kirchenbauer2023watermark}}  \\
        $h$ & Metric & $\temp:$ 0.8 & 0.9 & 1.0 & 1.1 & $\delta:$ 1.0 & 2.0 & 3.0 & 4.0 \\
        \cmidrule(rr){3-6} \cmidrule(rr){7-10}
        \multirow{2}{*}{$0$} 
            & S-BERT    & 0.60 & 0.56 & 0.52 & 0.44 & 0.63 & 0.61 & 0.57 & 0.50 \\
            & TPR       & 0.20 & 0.31 & 0.42 & 0.51 & 0.00 & 0.16 & 0.58 & 0.70 \\
            & TPR aug.  & 0.04 & 0.06 & 0.09 & 0.10 & 0.00 & 0.02 & 0.20 & 0.39 \\[4pt]
        \multirow{2}{*}{$1$} 
            & S-BERT    & 0.62 & 0.61 & 0.59 & 0.55 & 0.63 & 0.62 & 0.60 & 0.56 \\
            & TPR       & 0.35 & 0.51 & 0.66 & 0.77 & 0.02 & 0.41 & 0.77 & 0.88 \\
            & TPR aug.  & 0.04 & 0.10 & 0.20 & 0.36 & 0.00 & 0.05 & 0.30 & 0.58 \\[4pt]
        \multirow{2}{*}{$4$} 
            & S-BERT    & 0.62 & 0.62 & 0.61 & 0.59 & 0.62 & 0.62 & 0.60 & 0.57 \\
            & TPR       & 0.43 & 0.59 & 0.71 & 0.80 & 0.02 & 0.44 & 0.76 & 0.88 \\
            & TPR aug.  & 0.01 & 0.02 & 0.06 & 0.18 & 0.00 & 0.00 & 0.03 & 0.14 \\
        \bottomrule
        \end{tabular}
    }\vspace*{-0.3cm}
\end{table}

\subsection{Impact of Watermarks on Free-Form Generation Tasks}
Previous studies measure the impact on quality using distortion metrics such as perplexity or similarity score as done in Tab.~\ref{tab:robustness}.
However, such metrics are not informative of the utility of the model for downstream tasks~\cite{holtzman2019curious}, where the real interest of LLMs lies. 
Indeed, watermarking LLMs could be harmful for tasks that require very precise answers.
This section rather quantifies the impact on typical NLP benchmarks, in order to assess the practicality of watermarking.

LLMs are typically evaluated either by comparing samples of plain generation to a set of target references (free-form generation) or by comparing the likelihood of a predefined set of options in a multiple choice question fashion. 
The latter makes little sense in the case of watermarking, which only affects sampling.
We therefore limit our evaluations to free-form generation tasks.
We use the evaluation setup of LLaMA:
1) Closed-book Question Answering (Natural Questions~\cite{kwiatkowski2019natural}, TriviaQA~\cite{joshi2017triviaqa}): we report the $5$-shot exact match performance;
2) Mathematical reasoning (MathQA~\cite{hendrycks2021measuring}, GSM8k~\cite{cobbe2021training}), we report exact match performance without majority voting;
3) Code generation (HumanEval~\cite{chen2021Evaluating}, MBPP~\cite{austin2021program}), we report the pass@1 scores.
For \cite{kirchenbauer2023watermark}, we shift logits with $\delta=1.0$ before greedy decoding.
For \cite{aaronson2023watermarking}, we apply top-p at $0.95$ to the probability vector, then apply the watermarked sampling.

Tab.~\ref{tab:bench} reports the performance of LLaMA models on the aforementioned benchmarks, with and without the watermark and for different window size $h$. 
The performance of the LLM is not significantly affected by watermarking. 
The approach of Kirchenbauer~\etal (\ref{sec:ModDistri}) is slightly more harmful than the one of Aaronson~\etal (\ref{sec:ModSamp}), but the difference w.r.t. the vanilla model is small.
Interestingly, this difference decreases as the size of the model increases: models with higher generation capabilities are less affected by watermarking. A possible explanation is that the global distribution of the larger models is better and thus more robust to small perturbations.
Overall, evaluating on downstream tasks points out that watermarking may introduce factual errors that are not well captured by perplexity or similarity scores.

\section{Advanced Detection Schemes}
This section introduces improvements to the detection schemes of Sect.~\ref{sec:stats}.
Namely, it develops a statistical test when access to the LLM is granted, as well as multi-bit decoding.

\subsection{Neyman-Pearson and Simplified Score Function} 
The following is specific for the scheme of Aaronson~\etal~\cite{aaronson2023watermarking} (a similar work may be conducted with~\cite{kirchenbauer2023reliability}).
Under $\H_0$, we have $\vec{r}_v\sim\mathcal{U}_{[0,1]}$, whereas $\vec{r}_v\sim Beta(1/p_v,1)$ under $\H_1$ (see Corollary~\eqref{eq:Coro} in App.~\ref{app:aaronson_prob}). 
The optimal Neyman-Pearson score function is thus:
\begin{equation*}
    S_T = \sum_{t=1}^{T} \ln\frac{f_{\H_1}(\vec{r}_{x^{(t)}})}{f_{\H_0}(\vec{r}_{x^{(t)}})} = \sum_{t=1}^T \left(\frac{1}{\vec{p}_{x^{(t)}}}-1\right)\ln(\vec{r}_{x^{(t)}})+A
\end{equation*}
where $A$ doesn't depend on $\vec{r}$ and can thus be discarded. There are two drawbacks: (1) detection needs the LLM to compute $\vec{p}_{x^{(t)}}$, (2) there is no close-form formula for the p-value.  

This last point may be fixed by resorting to a Chernoff bound, yet without guarantee on its tightness:
$\text{p-value}(s) \leq e^{\sum_t \ln\frac{\lambda_t}{\lambda_t + c} -cs}$,
with $c$ solution of $\sum_t (c+\lambda_t)^{-1}=-s$ and $\lambda_t = p_{x^{(t)}} / (1-p_{x^{(t)}})$.
Experiments show that this detection yields extremely low p-values for watermarked text, but they are fragile: any attack increases them to the level of the original detection scheme~\eqref{eq:ScoreAaronson}, or even higher because generated logits are sensitive to the overall LLM context. 
An alternative is to remove weighting:
\vspace*{-0.4cm}
\begin{equation}
 S_T = \sum_{t=1}^T \ln\left(\vec{r}_{x^{(t)}}\right),
 \label{eq:Detection2}
 \vspace*{-0.2cm}
\end{equation}
whose p-value is given by: $\text{p-value}(s) = \frac{\gamma(T,-s)}{\Gamma(T)}$.
In our experiments, this score function does not match the original detection presented in~\cite{aaronson2023watermarking}.

\subsection{Multi-bit Watermarking}

\setlength{\textfloatsep}{0.2cm}
\begin{algorithm}[t] \small
    \caption{Multi-bit watermarking for LLMs}
    \label{alg:multi-bit}
    \begin{algorithmic}
        \State\hspace*{-0.3cm} \textbf{Requires}: {model LLM, secret's dimension $d = \textrm{max}(M, |\V|)$, watermark context width $h$}, message $m\in\{0,\ldots,M-1\}$
       
        \vspace*{0.2cm}\State\hspace*{-0.3cm} \emph{Generation (one step)}:
        \State logits $\vec{\boldsymbol\ell} \gets \text{LLM} \left( x^{(-C)},\dots, x^{(-1)} \right)$
        \State seed $\gets \mathsf{Hash}(x^{(-h)},\dots, x^{(-1)})$
        \State $\vec{r} \gets \mathsf{RNG_{seed}}(d)$
        \State $\vec{r}(m) \gets \mathsf{CyclicShift}(\vec{r},m) = \left( \vec{r}_m,..,\vec{r}_{d}, \vec{r}_{0}, ..,  \vec{r}_{m-1}  \right)$
        \State $x^{(0)} \gets \mathsf{Sample}(\vec{\boldsymbol\ell},\vec{r}(m)_{1,\dots,|\V|})$
    
        \vspace*{0.2cm}\State\hspace*{-0.3cm} \emph{Identification}:
        \State $\vec{S} \gets \vec{0}_d$
        \State{\textbf{for} $t \in \{ h, \dots, T\}$:}
            \State \quad seed $\gets \mathsf{Hash}(x^{(t-h)},\dots, x^{(t-1)})$
            \State \quad $\vec{r}^{(t)} \gets \mathsf{RNG_{seed}}(d)$
            \State \quad $\vec{S} \gets \vec{S} +  \mathsf{CyclicShift}(f(\vec{r}^{(t)}),x^{(t)})$
        \State $\vec{p} \gets \textrm{p-value}(\vec{S}_{1,\dots,M})$
        \State $m \gets \textrm{argmin}({\vec{p}}) $
        \State $p \gets 1 - (1 - \vec{p}_m)^M$
    \end{algorithmic}
\end{algorithm}

\subsubsection{Theory}
It is rather easy to turn a zero-bit watermarking scheme into multi-bit watermarking, by associating a secret key per message. 
The decoding runs detection with every key and the decoded message is the one associated to the key giving the lowest p-value $p$. 
The global p-value becomes $1-(1-p)^M$, where $M$ is the number of possible messages.

Running detection for $M$ keys is costly, since it requires $M$ generations of the secret vector.
This is solved by imposing that the secret vectors of the messages $m\in\{0,\ldots,M-1\}$ are crafted as circular shifts of $m$ indices of $\vec{r}=\vec{r}(0)$:
\begin{align*}\vspace{-0.5cm}
\vec{r}(m) &= \mathsf{CyclicShift}(\vec{r},m) \\
    &= \left( \vec{r}_m, \vec{r}_{m+1}, ..,\vec{r}_{d}, \vec{r}_{0}, ..,  \vec{r}_{m-1}  \right).
\vspace{-0.5cm}
\end{align*}
Generating $\vec{r}$ as a $d$-dimensional vector, with $d\geq|\V|$, we are able to embed $M\leq d$ different messages, by keeping only the first $|\V|$ dimensions of each circularly-shifted vector. 
Thus, the number of messages may exceed the size of the token vocabulary $|\V|$.

Besides, the scoring functions~\eqref{eq:ScoreKirchenbauer}~\eqref{eq:ScoreAaronson}
may be rewritten as:
\vspace*{-0.2cm}
\begin{equation}
S_T(m) = \sum_{t=1}^T f\left(\vec{r}^{(t)}(m)\right)_{x^{(t)}}  ,
\vspace*{-0.2cm}
\end{equation}
where $f: \R^d \mapsto \R^d$ is a component-wise function, and $x^{(t)}$ is the selected token during detection. 
This represents the selection of $f\left(\vec{r}^{(t)}(m)\right)$ at position $x^{(t)}$.
From another point of view, if we shift $f\left(\vec{r}^{(t)}\right)$ by $x^{(t)}$, the score for $m=0$ would be its first component, $m=1$ its second one, etc.
We may also write:
\vspace*{-0.4cm}
\begin{equation}
\vec{S}_T = \sum_{t=1}^T \mathsf{CyclicShift}\left( f\left(\vec{r}^{(t)}\right), x^{(t)} \right) ,
\label{eq:DetectionMultibit}
\vspace*{-0.2cm}
\end{equation}
and the first $M$ components of $\vec{S}_T$ are the scores for each $m$.\\
As a side note, this is a particular case of the parallel computations introduced by Kalker~\etal~\cite{JAWS}.

\vspace*{0.1cm}
\subsubsection{Experiments}
In a tracing scenario the message is the identifier of a user or a version of the model.
The goal is to decide if any user or model generated a given text (detection) and if so, which one (identification).
There are 3 types of error: \emph{false positive}: flag a vanilla text; \emph{false negative}: miss a watermarked text; \emph{false accusation}: flag a watermarked text but select the wrong identifier.

We simulate $M'$=$1000$ users that generate $100$ watermarked texts each, using the Guanaco-7b model. 
Accuracy can then be extrapolated beyond the $M'$ identifiers by adding identifiers with no associated text, for a total of $M>M'$ users.
Text generation uses nucleus sampling with top-p at $0.95$.
For~\cite{kirchenbauer2023watermark}, we use $\delta=3.0$, $\gamma=1/4$ with temperature $\theta$ at $0.8$.
For~\cite{aaronson2023watermarking}, we use $\theta = 1.0$.
For both, the context width is $h=4$.
A text is deemed watermarked if the score is above a threshold set for a given \emph{global} FPR (see~\ref{sec:stats}).
Then, the source is identified as the user with the lowest p-value.

Tab.~\ref{tab:identification} shows that watermarking enables identification because its performance is dissuasive enough. 
For example, among $10^5$ users, we successfully identify the source of a watermarked text 50\% of the time while maintaining an FPR of $10^{-6}$ (as long as the text is not attacked).
At this scale, the false accusation rate is zero (no wrong identification once we flag a generated text) because the threshold is set high to avoid FPs, making false accusations unlikely. 
The identification accuracy decreases when $M$ increases, because the threshold required to avoid FPs gets higher.
In a nutshell, by giving the possibility to encode several messages, we trade some accuracy of detection against the ability to identify users.

\begin{table}[t]
    \caption{Identification accuracy for tracing users by watermarking. 
    Sequences are between $4$ and $252$ tokens long, and $149$ on average.
    }
    \label{tab:identification}
    \centering
    \begin{tabular}{cr cccc}
    \toprule
    & Number of users $M$ & $10$ & $10^2$ & $10^3$ & $10^4$ \\ \midrule
    \multirow{2}{*}{FPR$=10^{-3}$} & Aaronson~\etal\cite{aaronson2023watermarking}      & 0.80 & 0.72 & 0.67 & 0.62 \\
    & Kirchenbauer~\etal\cite{kirchenbauer2023watermark}  & 0.84 & 0.77 & 0.73 & 0.68 \\ \midrule
    \multirow{2}{*}{FPR$=10^{-6}$} & Aaronson~\etal\cite{aaronson2023watermarking}	      & 0.61 & 0.56 & 0.51 & 0.46 \\
    & Kirchenbauer~\etal\cite{kirchenbauer2023watermark} 	                              & 0.69 & 0.64 & 0.59 & 0.55 \\
    \bottomrule
    \end{tabular}
\end{table}

\section{Conclusion}

This research offers theoretical and empirical insights that were kept aside from the literature on watermarks for LLMs.
Namely, existing methods resort to statistical tests which are biased, delivering incorrect false positive rates.
This is fixed with grounded statistical tests and a revised scoring strategy.
We additionally introduce evaluation setups, and detection schemes to consolidate watermarks for LLMs.
Further work may investigate how to adapt watermarks for more complex sampling schemes (\eg beam search as in \cite{kirchenbauer2023watermark}), since generation yield significantly better quality with these methods.

Overall, we believe that watermarking is both reliable and practical. 
It already holds many promises as a technique for identifying and tracing LLM outputs, while being relatively new in the context of generative models. 

\section*{Acknowledgments}
Work supported by ANR / AID under Chaire SAIDA ANR-20-CHIA-0011.
We also thank Thomas Scialom, Hervé Jégou and Matthijs Douze for their insights throughout this work.

\bibliography{references}
\bibliographystyle{ieeetr}

\clearpage
\onecolumn
\appendix

\subsection{Demonstrations for \texorpdfstring{\cite{aaronson2023watermarking}}{}}\label{app:aaronson_demo}

\emph{1) Sampling probability}\vspace{0.2cm}
\label{app:aaronson_prob}

\noindent\emph{Proposition.}
Consider a discrete distribution $\vec{p}=(p_1,\ldots,p_V)$
and $V$ random variables $\vec{R} = (R_1,\ldots,R_V)$ s.t. $R_v\overset{iid}{\sim}\mathcal{U}_{[0,1]}$. 
Let $V^\star = \arg \max_v R_v^{1/p_v}$.
Then $\Prob(V^\star=v) = p_v$.\\

\noindent
\emph{Proof.}
For any $v \in \V$, $R_v\overset{iid}{\sim}\mathcal{U}_{[0,1]}$ so, $- \ln(R_v)$ follows an exponential distribution $\mathcal{E}(1)$.
Let $Z_v := -\frac{1}{p_v} \ln(R_v)$. By construction, $Z_v\sim\mathcal{E}(p_v)$, with density $f_{Z_v}(z) = p_v e^{-p_v.z}$.
We now have:
\begin{equation}
V^\star = \arg \max_v R_v^{\frac{1}{p_v}} = \arg \min_v Z_v.
\end{equation}
A well known result about exponential laws is that (see \href{https://francisbach.com/the-gumbel-trick/}{the-gumbel-trick} for following lines):
\begin{eqnarray}
\underline{Z}  &=& \min_v Z_v \sim \mathcal{E}\left(\sum_v p_v\right)=\mathcal{E}\left(1\right),\\ \label{eq:sampling}
\Prob(V^\star=v) &=& \frac{p_v}{\sum_j p_j}  = p_v.
\end{eqnarray}

This shows that for a given secret vector $\vec{r}$, the watermarking chooses a word which may be unlikely (low probability $p_{V^\star}$). 
Yet, on expectation over the secret keys, ie. over r.v. $\vec{R} = (R_1, \ldots, R_V)$, the distribution of the chosen token follows the distribution given by the LLM.\\

\noindent\emph{Corollary.} $R_{V^\star} \sim Beta(1/p_{V^\star}, 1)$.\\

\noindent
\emph{Proof.}
\begin{equation}
\underline{Z}  = Z_{V^\star} = -\frac{1}{p_{V^\star}} \ln(R_{V^\star}) \sim \mathcal{E}(1),
\label{eq:Coro}
\end{equation}
which translates to $R_{V^\star} = e^{-p_{V^\star} E}$ with $E\sim\mathcal{E}(1)$, with p.d.f. $f_{R_{V^\star}}(r) = \frac{r^{\frac{1}{p_{V^\star}}-1}}{p_{V^\star}}$. 
Therefore, $R_{V^\star} \sim Beta(1/p_{V^\star}, 1)$.

\vspace{0.4cm}
\emph{2) Detection}\vspace{0.1cm}
\label{app:aaronson_pvalue}

\noindent
We denote by $x^{(1)}, \ldots, x^{(T)}$ the sequence of tokens in the text, 
by $\vec{p}^{(t)}$ the probability vector output by the LLM and by $\vec{R}^{(t)} \in [0,1]^{|\mathcal{V}|}$ the key random vector at time-step $t$.
We define $R_t := R^{(t)}_{x^{(t)}}$ and $p_t := p^{(t)}_{x^{(t)}}$ at time-step $t$.
The score is $S_T=-\sum_{t=1}^{T} \ln (1-R_t)$.
\\ \noindent \\ 
\noindent\emph{Proposition ($p$-value under $\H_0$).}
The $p$-value associated to a score $s$ is defined as:
\begin{equation}
\text{$p$-value}(s) = \Prob(S_T>s | \H_0) = \frac{\Gamma(T,s)}{\Gamma(T)},
\end{equation}
where $\Gamma(T,s)$ is the \emph{upper} incomplete gamma function.\\

\noindent
\emph{Proof.}
Under $\H_0$, the assumption is s.t. $R_t\overset{iid}{\sim}\mathcal{U}_{[0,1]}$. 
Then, $- \ln(1-R_t)$ follows an exponential distribution $\mathcal{E}(1)$.
Therefore $S\sim\Gamma(T,1)$ (see \href{https://en.wikipedia.org/wiki/Gamma_distribution#Summation}{sum of Gamma distributions}). Therefore the $p$-value associated to a score $s$ is 
\begin{equation}
    \text{$p$-value}(s) = 1 - \frac{\gamma(T,s)}{\Gamma(T)} = \frac{\Gamma(T, s)}{\Gamma(T)} ,
\end{equation}
where $\Gamma(T,s)$ is the \emph{upper} incomplete gamma function, $\gamma(T,s)$ is the \emph{lower} \href{https://en.wikipedia.org/wiki/Incomplete_gamma_function}{incomplete gamma function}. \\

\noindent\emph{Corollary.} Per token, 
\begin{equation}
\mu_0 = \E(S_T/T|\H_0) = 1,\quad \sigma_0^2 = \mathbb{V}(S_T/T|\H_0) = 1/T.
\end{equation}

\vspace{0.3cm}
\noindent\emph{Proposition (Bound on expected score under $\H_1$).}
Under $\H_1$, 
$\displaystyle \mathbb{E}(S_T) \geq T +  \left( \frac{\pi^2}{6} -1 \right) H_T $, 
where $H_T = - \sum_{t=1}^T p_t\ln(p_t)$ is the entropy of the completion.\\

\noindent
\emph{Proof.}
From~\eqref{eq:Coro}, $R_t=\exp(-p_t E)$ with $E\sim \mathcal{E}(1)$, so:
\begin{align*}
    \mathbb{E}(S) &= - \mathbb{E} \left[ \sum_{t=1}^T \ln (1-\exp(-p_t E)) \right] \\
    &= - \sum_{t=1}^T \int_0^\infty \ln (1-e^{-p_t x}) e^{-x} dx \\
    &= - \sum_{t=1}^T \int_0^1 \frac{1}{p_t} r^{1/p_t-1} (-\ln ( 1 - r)) dr  \\ 
    & \text{ \qquad (by change of variable $x = -1/p_t \ln (r) $ )} 
\end{align*}
Then, using integration by parts with $u = 1 - r^{1/p_t}$ and $v = \ln(1-r)$, the integral becomes:
\begin{align*}
    -\int_0^1 \frac{1}{p_t} r^{1/p_t-1} \ln ( 1 - r) dr &= \int_0^1 \frac{1-r^{1/p_t}}{1-r} dr = \H_{1/p_t}
\end{align*}
where $\H_{z}$ is the $z$-th \href{https://en.wikipedia.org/wiki/Harmonic_number}{harmonic number}
also defined as $\H_{z} = \sum_{n=1}^\infty \frac{1}{n} - \frac{1}{n+z}$.
Therefore, we have:
\begin{align*}
    -\int_0^1 \frac{1}{p_t} r^{1/p_t-1} \ln ( 1 - r) dr &= 
        \sum_{n=1}^\infty \frac{1}{n} - \frac{1}{n+1/p_t} \\
    &= 1 + \sum_{n=1}^\infty \frac{1}{n+1} - \frac{1}{n+1/p_t}.
\end{align*}
Now, $\forall n\in \mathbb{N^\star}$, we have:
\begin{align*}
   (n+1)^2 \left(\frac{1}{n+1} - \frac{1}{n+1/p_t}\right) &= \frac{(n+1)(n+1/p_t) - (n+1)^2}{n + 1/p_t} \\
    &=  \frac{1+n}{1/p_t + n} \left( 1/p_t -1\right) \\
    &\geq -  \frac{1+n}{1/p_t + n} \ln(p_t) \\
    &\geq -  \, p_t \ln(p_t).
\end{align*}
Therefore, by summing over all $t\in [1,T]$,
\begin{align*}
    \mathbb{E}(S) &\geq T +  \left(\sum_{n=1}^\infty \frac{1}{(n+1)^2}\right)\left(\sum_{t=1}^T- p_t\ln(p_t) \right) \\
    &=T +  \left( \frac{\pi^2}{6} -1 \right) H_T.
\end{align*}
\\ \noindent \\
\noindent\emph{Proposition (Variance of score under $\H_1$).}
$\displaystyle\mathbb{V}(S_T)\leq T\frac{\pi^2}{6}$. \\

\noindent
\emph{Proof.}
For $R_{t}\sim Beta(1/p_t, 1)$:
\begin{equation}
\mathbb{V}(\ln(1-R_t)) = \psi_1(1) - \psi_1(1+1/p_t)
\end{equation}
where $\psi_1$ is the trigamma function, which can be expressed as the following serie $\psi_1(z) = \sum_{n=0}^{\infty} 1/(n+z)^2$. 
Then $\psi_1(1) = \pi^2/6$ and $\psi_1(1+1/p_t)>0$, so that $\mathbb{V}(\ln(1-R_t)) \leq \pi^2/6$.
The results comes because the sampled tokens are independent. \\

\subsection{Free-form evaluations}
We provide in Table~\ref{tab:bench-full} the full results of the free-form evaluations of the different models.
This extends the results of Table~\ref{tab:bench} in the main paper.
The models are evaluated with the same evaluation protocol as in LLaMA.

\begin{table*}
    \centering
    \caption{ }
    \label{tab:bench-full}
    \begin{tabular}{lllrrrrrrr}
    \toprule
     &  &  & GSM8K & Human Eval & MathQA & MBPP & NQ & TQA & Average \\
    Model & WM Method & h &  &  &  &  &  &  &  \\
    \midrule
     \multirow[t]{15}{*}{7B} 
     & None & - & 10.31 & 12.80 & 2.96 & 18.00 & 21.72 & 56.89 & 20.45 \\
     \cmidrule{2-10} 
     & Aaronson \etal & 0 & 10.54 & 12.80 & 3.00 & 18.00 & 21.77 & 56.88 & 20.50 \\
      &  & 1 & 10.31 & 12.80 & 2.88 & 18.20 & 21.75 & 56.87 & 20.47 \\
      &  & 2 & 10.31 & 12.80 & 2.94 & 18.00 & 21.75 & 56.86 & 20.44 \\
      &  & 3 & 10.39 & 12.80 & 2.96 & 18.20 & 21.69 & 56.85 & 20.48 \\
      &  & 4 & 10.39 & 12.80 & 2.98 & 17.80 & 21.80 & 56.88 & 20.44 \\
      &  & 6 & 10.61 & 12.80 & 2.96 & 18.00 & 21.75 & 56.86 & 20.50 \\
      &  & 8 & 10.46 & 12.80 & 2.90 & 18.20 & 21.75 & 56.85 & 20.49 \\
      \cmidrule{2-10} 
      & Kirchenbauer \etal & 0 & 9.63 & 12.80 & 2.20 & 16.20 & 20.06 & 55.09 & 19.33 \\
      &  & 1 & 11.14 & 9.76 & 2.82 & 16.00 & 19.50 & 55.30 & 19.09 \\
      &  & 2 & 11.07 & 6.71 & 2.62 & 16.00 & 20.44 & 55.07 & 18.65 \\
      &  & 3 & 10.16 & 10.98 & 2.38 & 14.40 & 20.08 & 55.65 & 18.94 \\
      &  & 4 & 10.77 & 9.15 & 2.76 & 16.40 & 20.17 & 55.14 & 19.06 \\
      &  & 6 & 10.01 & 9.76 & 3.16 & 17.00 & 20.78 & 54.90 & 19.27 \\
      &  & 8 & 11.37 & 11.59 & 2.90 & 16.40 & 20.66 & 55.36 & 19.71 \\
    \midrule
    \multirow[t]{15}{*}{13B} 
    & None & - & 17.21 & 15.24 & 4.30 & 23.00 & 28.17 & 63.60 & 25.25 \\
    \cmidrule{2-10} 
    & Aaronson \etal & 0 & 17.29 & 15.24 & 4.24 & 22.80 & 28.17 & 63.60 & 25.22 \\
     &  & 1 & 17.21 & 15.24 & 4.30 & 22.80 & 28.20 & 63.61 & 25.23 \\
     &  & 2 & 17.51 & 15.24 & 4.20 & 22.80 & 28.20 & 63.59 & 25.26 \\
     &  & 3 & 17.44 & 15.24 & 4.22 & 22.60 & 28.23 & 63.57 & 25.22 \\
     &  & 4 & 17.21 & 15.24 & 4.20 & 22.60 & 28.20 & 63.63 & 25.18 \\
     &  & 6 & 16.98 & 15.24 & 4.28 & 23.20 & 28.23 & 63.61 & 25.26 \\
     &  & 8 & 17.21 & 15.24 & 4.22 & 22.80 & 28.20 & 63.62 & 25.22 \\
     \cmidrule{2-10} 
     & Kirchenbauer \etal & 0 & 14.33 & 14.02 & 3.04 & 20.80 & 24.32 & 62.13 & 23.11 \\
     &  & 1 & 17.29 & 14.63 & 3.62 & 21.20 & 25.12 & 62.23 & 24.02 \\
     &  & 2 & 16.45 & 11.59 & 3.54 & 20.60 & 25.54 & 62.44 & 23.36 \\
     &  & 3 & 17.06 & 16.46 & 3.58 & 19.80 & 25.90 & 62.37 & 24.20 \\
     &  & 4 & 16.76 & 15.85 & 4.08 & 21.20 & 24.49 & 62.24 & 24.10 \\
     &  & 6 & 15.85 & 14.63 & 4.00 & 18.20 & 26.32 & 62.19 & 23.53 \\
     &  & 8 & 17.29 & 14.63 & 3.68 & 21.00 & 25.46 & 62.17 & 24.04 \\
     \midrule
    \multirow[t]{14}{*}{30B} 
     & None & - & 35.10 & 20.12 & 6.80 & 29.80 & 33.55 & 70.00 & 32.56 \\
     \cmidrule{2-10} 
     & Aaronson \etal & 0 & 35.48 & 20.12 & 6.88 & 29.80 & 33.52 & 69.98 & 32.63 \\
     & & 1 & 35.33 & 20.73 & 6.88 & 29.60 & 33.52 & 70.03 & 32.68 \\
     & & 2 & 35.33 & 20.73 & 6.94 & 30.00 & 33.49 & 70.00 & 32.75 \\
     & & 3 & 35.71 & 20.73 & 6.92 & 30.00 & 33.52 & 70.02 & 32.82 \\
     & & 4 & 35.10 & 20.12 & 6.90 & 29.80 & 33.49 & 70.01 & 32.57 \\
     & & 6 & 35.33 & 20.73 & 6.86 & 29.80 & 33.49 & 69.98 & 32.70 \\
     & & 8 & 35.33 & 20.73 & 6.94 & 30.00 & 33.52 & 70.01 & 32.75 \\
     \cmidrule{2-10} 
     & Kirchenbauer \etal & 0 & 31.84 & 21.95 & 6.88 & 28.40 & 31.66 & 69.03 & 31.63 \\
     &  & 1 & 35.56 & 20.73 & 7.54 & 28.80 & 31.58 & 68.98 & 32.20 \\
     &  & 2 & 33.21 & 17.07 & 6.48 & 27.40 & 31.83 & 69.44 & 30.91 \\
     &  & 3 & 33.89 & 24.39 & 6.54 & 27.80 & 32.49 & 69.22 & 32.39 \\
     &  & 4 & 34.12 & 22.56 & 6.96 & 28.80 & 31.55 & 68.74 & 32.12 \\
     &  & 6 & 34.34 & 24.39 & 7.32 & 29.80 & 31.63 & 69.08 & 32.76 \\
     &  & 8 & 34.95 & 20.12 & 7.42 & 27.20 & 32.08 & 69.31 & 31.85 \\
    \bottomrule
    \end{tabular}
\end{table*}

\end{document}